\documentclass[review]{elsarticle}
\usepackage{lineno,hyperref}
\usepackage{amsmath,amssymb,amsfonts}
\usepackage{graphicx}
\usepackage{natbib}

\begin{document}

\begin{frontmatter}

\title{Integration of Autoencoder and Functional Link Artificial Neural Network for Multi-label Classification}

\author[1]{Anwesha Law}
\ead{anweshalaw_r@isical.ac.in}


\author[1]{Ashish Ghosh\corref{correspondingauthor}}
\cortext[correspondingauthor]{Corresponding author}
\ead{ash@isical.ac.in}

\address[1]{Machine Intelligence Unit, Indian Statistical Institute, Kolkata, India}

%

\begin{abstract}
Multi-label (ML) classification is an actively researched topic currently, which deals with convoluted and overlapping boundaries that arise due to several labels being active for a particular data instance. We propose a classifier capable of extracting underlying features and introducing non-linearity to the data to handle the complex decision boundaries. A novel neural network model has been developed where the input features are subjected to two transformations adapted from multi-label functional link artificial neural network and autoencoders. First, a functional expansion of the original features are made using basis functions. This is followed by an autoencoder-aided transformation and reduction on the expanded features. This network is capable of improving separability for the multi-label data owing to the two-layer transformation while reducing the expanded feature space to a more manageable amount. This balances the input dimension which leads to a better classification performance even for a limited amount of data. The proposed network has been validated on five ML datasets which shows its superior performance in comparison with six well-established ML classifiers. Furthermore, a single-label variation of the proposed network has also been formulated simultaneously and tested on four relevant datasets against three existing classifiers to establish its effectiveness.
\end{abstract}


\end{frontmatter}


\section{Introduction}
Classification is one of the most popular topics in machine learning and is widely performed by various types of supervised-learning approaches.
When discussing about classification, the traditional approach is indicated by default. Here, each data instance is associated with a single class, which is thus termed as single-label classification. However, real world scenarios follow a different track of problems where data can be annotated with multiple labels at a time. This gave rise to a sub-domain known as multi-label (ML) classification \citep{herrera2016multilabel}. Real-life examples of multi-label data can be easily seen around us. Social media posts using various hashtags (labels) for a single image or text, movies categorized under different genres, and so on. In each of these cases, a set of labels is associated with a particular instance of the dataset which determines the classes associated with them. However, due to more classes linked with each data, there is an increase in complexity of decision space. The class boundaries are much more convoluted and overlapping due to the increased generality of ML classification. In a way, ML classification can be seen as a superset of single-label classification where the classifiers need to be able to output several predictions at once.

Keeping the complex nature of the problem in mind, there are a few approaches which researchers have been using to handle ML data. The existing classifiers present in literature can be divided into few groups: data transformation \citep{gonccalves2003preliminary,read2009classifier}, problem adaptation \citep{zhang2007ml,law2019multi} and ensemble classifiers \citep{read2009classifier,cheng2010bayes,tsoumakas2007random}. Among these, the problem adaptation techniques are the most convenient and are widely used.
Some benchmark problem adaptation methods include the use of support vector machines (SVM) \citep{elisseeff2002kernel}, multi-layer perceptron (MLP) \citep{zhang2006multilabel}, k-nearest neighbours \citep{zhang2007ml}, decision-trees \citep{clare2001knowledge} and probabilistic classifiers \citep{cheng2010bayes}. While dealing with the complex nature of ML classification, most of these methods handle the data as it is.
However, it is seen that neural networks (NNs) are quite capable of handling complexities without actually creating a bulky model. They are loosely inspired by the working of a human brain and are one of the most popular and widely used tool for machine learning. They inherently bring non-linearity by increasing the number of layers in the model which are able to follow disparate structures in the datasets. Although lot of work has been done in single-label classification using NN, it is to be noted that comparatively very few works include the usage of NNs for ML classification.

In this current work, the authors have attempted to construct a novel two-layer transformation network, adapted from multi-label functional link artificial neural network (MLFLANN), previously proposed by the authors in \citep{law2017functional}, and the well-known autoencoders (AE). This adaptation of MLFLANN and AE has been aptly named as the AutoEncoder integrated MLFLANN (AE-MLFLANN). This network is capable of overcoming few drawbacks faced in multi-label classification previously.
In the first layer of our network, we perform functional expansion of features inspired from MLFLANN. The main motivation to incorporate MLFLANN is its compact structure and complexity handling capability which seems suitable for multi-label classification. The input features are functionally expanded to a higher dimension, thus giving rise to a decision space with increased separability. This introduces non-linearity in the data, similar to that of multi-layer perceptrons. It is an attempt to improve the input space, thereby, increasing the convergence. However, the transformed data helps to improve the multi-label feature space only to some extent. The expansion of input space might not always give rise to an optimal representation. There is still some scope of further transforming the data which will lead to more improved performance. Also, the functional expansion leads to the increase in the input dimension, which poses a problem for multi-label data.
To handle both these issues, we introduce a second feature transformation-cum-reduction layer incorporating autoencoders. The second layer is created from the encoder section of an AE, which is capable of transforming the features to a comparatively reduced and improved space. Autoencoders are widely known to implicitly extract features for classification tasks, while successfully transforming the data \cite{law2019multi}. It is capable of generating a suitable representation through unsupervised learning.

To concisely describe the network, it can be said that, the input features are functionally expanded in the first layer. In the second layer, these expanded features are passed through an AE which generates a favourable representation in a reduced feature space. These reduced and transformed features are then mapped to the output layer. This AE-MLFLANN model is capable of good multi-label classification as it can handle the complex decision space better by transforming the data through two consecutive layers. The proposed work has highlighted the importance of the two transformation layers, which can be extended further to build deeper networks. This lies outside the scope of this paper and will be explored in future. AE-MLFLANN has been experimentally shown to perform better than six benchmark methods over five multi-label datasets.
As per the knowledge of the authors, the proposed neural network model does not exist in literature, hence its application has been explored from multi-label domain to traditional single-label domain as well. This single-label version of the proposed model, named AutoEncoder integrated single-label FLANN (AE-SLFLANN), has been separately tested on four relevant datasets to analyse its success.

The contribution of this work can be highlighted as follows.
\begin{itemize}
  \item Introducing a novel two-layer network based on MLFLANN and AE specifically for multi-label classification.
  \item Improving separability in multi-label data by first applying functional expansion layer, followed by additional transformation by autoencoder layer.
  \item The increased feature dimension caused by the first layer is reduced by consecutive AE in the second layer. This maintains a balance between the feature space and sample size, which leads to a good training of the classifier with limited data.
  \item Introducing the single-label variation of this novel two-layer network.
  \item Experimental analysis of both single-label and multi-label classification networks with benchmark datasets.
\end{itemize}

The rest of the paper is organized as follows. In Section \ref{Sec_Back} related works is discussed.
In Section \ref{Sec_prop} some background on multi-label classification followed by the proposed model, termed as AE-MLFLANN is described.
In Section \ref{Sec_exp}, results are put and Section \ref{Sec_conc} concludes the paper.

\section{Related Works}\label{Sec_Back}
The field of multi-label learning has been developing since the past decade. There are quite a few researchers who had started exploring this field and had taken different approaches to handle multi-label data. Multi-label classification strategies are broadly divided into two categories, namely, data transformation and problem adaptation.

\subsection{Data Transformation Methods}
Data transformation techniques involve converting the original ML data into one or more simpler datasets that can be delivered to traditional binary/single-label classifiers. In a certain way, these methods act as a preprocessing phase, producing new datasets from the original ones.
Binary relevance (BR) \citep{gonccalves2003preliminary}, creates $C$ number of independent classifiers, where $C$ is equal to the dimension of label space. Each classifier works on the same number of input instances in the original ML data, but each classifier $C_i$ positively labels the $i^{th}$ label and others as negative. Such a method succumbs to inconveniences like dismissing of label correlations and label imbalance.
Label powerset (LP) \citep{boutell2004learning} encodes each label set of an instance to a unique number thereby, converting the multi-label classification to single-label classification. Though, if the number of unique label sets are greater, then it might not output the correct result. LP set also doesn't provide any ranking information among labels for an instance.

Some ensemble approaches also exist in literature. Classifier chains (CC) \citep{read2009classifier}, is seen as an improvement over BR. CC also uses the same number of classifiers as in BR, i.e., same as the number of classes, but chosen at random order. The first classifier is trained on the input attributes, the output label obtained is added back to the input space, this new input space is used for second classifier and the process is continued.
In ensemble of classifier chains (ECC) \citep{read2009classifier}, several classifier chains are trained with random ordering of labels and subsets of training instances for ML classification.

\subsection{Problem Adaptation based Classifiers}
Problem adaptation based classifiers, contrary to data transformation methods, adapts the algorithm to the ML data.
One of the most popularly used algorithm is ML-KNN \citep{zhang2007ml}. It is a lazy learning technique and is a multi-label adaptation of the traditional k-nearest neighbors classifier. It primarily works on the principle of MAP (maximum a-posteriori).
\citep{schapire2000boostexter} proposed the BoosTexter approach that aims at combining an ensemble of weak classifiers to obtain a single strong classifier. The BoosTexter is a multi-label extension of the popular ensemble learning approach, AdaBoost.
\citep{clare2001knowledge} implements a C4.5 decision tree to perform multi-label gene expression data classification. It uses a modified definition for entropy.
One of the first neural network based model for ML classification is backpropagation based multi-label learning (BP-MLL) \citep{zhang2006multilabel}. It is a two-layer feed-forward network, trained using backpropagation with a cost function that incorporates the ranking of labels. At the end, a unique threshold for each testing instance is calculated based on the processing of training set, BP-MLL depends on the number of neurons used in the hidden layers and also a substantial amount of time is required in training the network.
Multi-label extreme learning machine (MLELM) with a single hidden layer has been applied for ML classification in \citep{sun2016extreme}, but due to its random initialization of weights and only one pass of training, the method becomes unsuitable for complex ML data.
Multi-label functional link artificial neural network (MLFLANN) \citep{law2017functional} is a classifier where the input features are functionally expanded in the first layer. It is a simplified network for classification but its single layer feed-forward network architecture limits its capability to classify convoluted ML data extremely well. Different combinations for its basis functions and learning mechanisms have been explored by the authors in \cite{law2019optimize} to determine the optimal combination experimentally.

From discussions in previous sections, it can be seen that there is a limited exploration involving usage of NNs for ML classification and the models that exist have some shortcomings which needs to be handled. The proposed work intends to improve upon the previous work on MLFLANN and handle the shortcomings posed by it and other existing works.

\section{Autoencoder integrated multi-label functional link artificial neural network} \label{Sec_prop}
In this article, a two-layer transformation based neural network has been proposed for multi-label classification which incorporates MLFLANN and autoencoders.
Although MLFLANN is a compact and interesting model, it has not been sufficiently explored in the domain of multi-label classification.
In the basic model devised in \citep{law2017functional}, the data undergoes a single-level functional expansion which improves the multi-label data separability to some extent.
However, the single transformation provided by the basis functions seems to improve the separability and thus the classification, only to a certain extent. This motivated us to take it a step further, and introduce another layer of transformation which can further improve the classification capability of the network. This led to the conception of a two-layer network with the second transformation credited to an autoencoder. It seemed suitable to use an AE in this scenario, since, our aim is to transform as well as reduce the number of features. Without having to manually select features, AEs are capable of generating good representation of the original data in a new space. Thus, the multi-label data in the proposed AE-MLFLANN undergoes two feature transformations and reduction followed by learning and final classification.
In this regard, few preliminaries and the proposed architecture have been discussed.

\subsection{Mathematical Representation of Multi-Label Classification}
While dealing with a multi-label classification scenario, the data representations vary slightly when compared to single-label classification.
The $i^{th}$ input instance, $X_i$ with $d$ features is represented as, $X_i = \{x_{i1}, x_{i2}, ..., x_{id}\}$. For each such instance, a corresponding label-set $Y_i$ of $C$ dimensions is defined as, $Y_i = \{y_{i1}, y_{i2}, ..., y_{iC}\}$.
In this label-set, each $y_{ik}$, where, $1 \leq k \leq C$ equals to $0$ if it is not relevant to the $i^{th}$ data instance and equals to $1$ when it is relevant. The task of any multi-label classifier is to determine the correct label-set of relevant and irrelevant labels for any unknown data instance.




\subsection{Proposed Architecture}
\begin{figure*}[htbp]
  \centering
  \includegraphics[width=10cm]{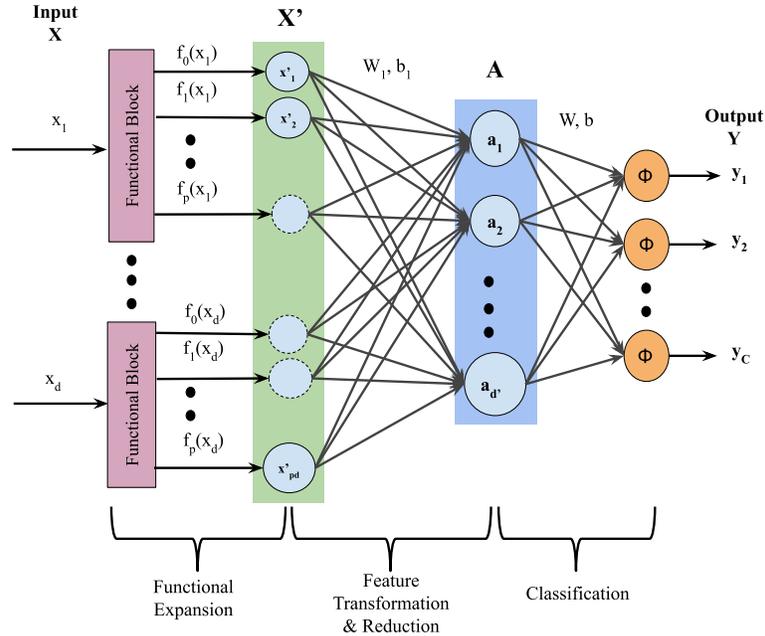}
  \caption{Proposed model AE-MLFLANN}\label{fig_aemlflann}
\end{figure*}

The proposed model employs autoencoders (AEs) in coalition with MLFLANN (AE-MLFLANN) to perform efficient multi-label classification.
The architecture of the proposed model has been shown in Fig. \ref{fig_aemlflann}. The model consists of three phases as follows.

\subsubsection{Functional Expansion}
The first layer is inspired by a multi-label functional link artificial neural network (MLFLANN). Neural networks have been the most popular models to solve complex classification tasks. However, due to the varying complexities of problems, number of hidden layers and neurons in the hidden layers need to be changed, giving rise to a more and more complex model. To overcome the bottlenecks that are associated with multi-layer perceptrons, higher order neural networks (HONNs) have been considered as an alternative. \citep{misra2007functional} demonstrated the use of FLANN, a type of HONN which is a flat network which requires no hidden layers for classification tasks. It also has the ability to achieve convergence faster with a lower training complexity. FLANN has proven to be very efficient in single-label classification \citep{dehuri2012improved}. It has also be adapted for multi-label classification by the authors in \cite{law2017functional}.

The first layer of AE-MLFLANN takes the original input and functionally expands it using some basis functions. The functional expansion acts on an element of a pattern or on the entire pattern by generating a set of linearly independent functions. Thereby, it effectively increases the dimension of the input and brings greater discriminating capability to the feature space. It comprises of $d$ input nodes corresponding to the original input features, $X=\{x_1, ..., x_d\}$, each of which are expanded using $p$ basis functions. This results in a $d \times p$ functionally expanded feature space. The input pattern $X$ is enhanced with the functional expansion block which results in the following pattern $X'$.


\begin{eqnarray}\label{Eq_MLFLANN}
\nonumber  X' &=& \{f_0(x_1), f_1(x_1), ..., f_p(x_1), \\
\nonumber & & \vdots \\
   & & f_0(x_d), f_1(x_d), ..., f_p(x_d)\}
\end{eqnarray}
Fig. \ref{fig_aemlflann} shows the functional blocks, which transform each feature, $x_i$, to its corresponding higher order representation. The functional expansion performed in the first layer can be done using various basis functions, like, trigonometric, Chebychev, power polynomials, etc.
Various such functional expansions have been experimented by the authors in \citep{law2019optimize}, some of which have proven to be fruitful for multi-label classification. Keeping that in mind, we have used trigonometric basis functions for functional expansion in the first layer.

The functional expansion provided by a MLFLANN is capable of transforming the data to a higher dimension, but in a restricted way.
The purpose of this expansion, is to bring about non-linearity in the data, which is otherwise impossible for a shallow neural network to achieve. However, this expansion alone can only introduce a restricted amount of separability in the complex decision space of multi-label data. If another level of transformation can be introduced in the data, it might be even more separable. To achieve this, an AE is to be used.

\subsubsection{Autoencoder Transformation and Feature Reduction}
Once the new set of features, $X'$, are obtained through functional expansion, they are then fed to an autoencoder. The AE brings about another level of implicit transformation in the network, which is capable of encoding the input features into a new set of features $A$.

Autoencoder is a popular variant of neural networks which is widely used in research. The aim of using an autoencoder is to learn an efficient data representation or encoding in an unsupervised manner, typically for dimensionality reduction. The simplest form of an autoencoder is similar to a MLP model, but has a symmetrical encoding and decoding architecture.
\begin{figure}[htbp]
  \centering
  \includegraphics[width=10cm]{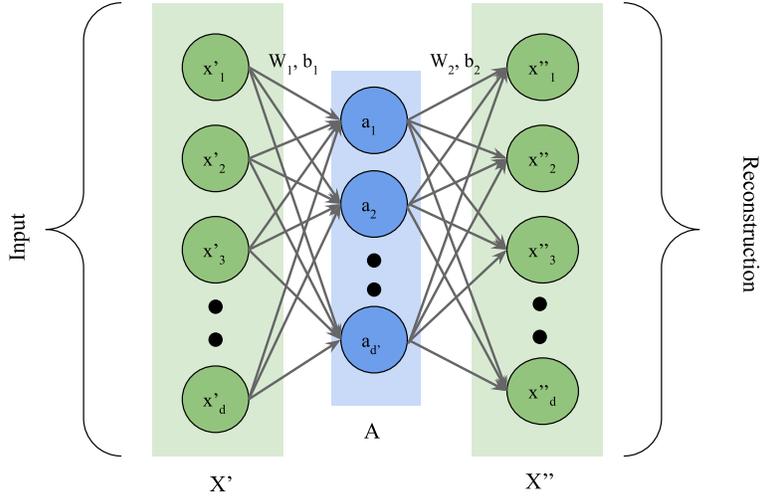}
  \caption{Single-layer under-complete autoencoder}
  \label{Fig_AE}
\end{figure}
Fig. \ref{Fig_AE} illustrates the architecture of a single intermediate layer autoencoder, with the first layer having $d$ nodes and $X'$ being the input. The last layer is the reconstructed input layer $X'' \in \mathbb{R}^d$. The AE is trained to reconstruct the input as closely as possible. The intermediate layer learns an encoded latent representation of the input in the course of its training. It basically performs the task of an encoder, whereas, the decoding happens from the intermediate layer to the output layer. The aim of such a network as mentioned earlier is to minimize the error between input vector and the output vector.
If the number of neurons in the encoding layers are lesser than the original number of features, the autoencoder is said to be under-complete and if it is greater than the input it is said to be over-complete. It manipulates the dimension of the encoded features, thus leading to dimensionality reduction or expansion.
The working of the AE is quite similar to that of a multi-layer perceptron. The forward propagation which maps the input vector $X'$ to the latent representation $A \in \mathbb{R}^{d'}$ and then maps $A$ to the output $X''$ (reconstruction) can be formulated as:
\begin{align}
    A \ &= \sigma_1(W_1 \cdot X' + b_1)\\
    X'' &= \sigma_2(W_2 \cdot A + b_2)
\end{align}
Here, $W_1$ and $b_1$ represents the weight matrix and bias from input to hidden layer and $W_2$ and $b_2$ represents the weight matrix from hidden layer to reconstruction layer. $\sigma_1, \sigma_2$ represents the activation functions which can be chosen from sigmoid activation, rectified linear unit (ReLU), hyperbolic tangent function, etc.
The AE iteratively modifies the weights in the network using backpropagation, to finally learn to reconstruct the original input in the output layer.

In the AE-MLFLANN model (Fig. \ref{fig_aemlflann}), the autoencoder transformation is incorporated by the use of the encoder weights, $W_1, b_1$, from the trained AE (Fig. \ref{Fig_AE}). These weights are for the connections between the first transformation layer to the second. $X'$ can be now modified to $A$ by using the autoencoder weights, just like it was done in the standalone autoencoder.

Along with feature space transformation, dimensionality reduction is also performed with the help of the AE. In general, AEs are known to extract good features from data. If the size of the intermediate layer of the AE is varied, it might result in feature expansion or reduction. However, in the proposed scenario, to handle the already expanded feature space by the basis functions, we opt for feature reduction by the AE. This helps with the additional bottleneck of increased input dimensionality in the previous layer. This was a drawback of MLFLANN which is being overcome in AE-MLFLANN. This feature encoding reduces the size of the network at this stage, thus in turn, having lesser weights to learn.

Once the data has been transformed completely, the encoded and reduced representation $A=\{a_1, a_2, ..., a_{d'}\}$, is further connected to the output layer of the network.
These $d'$ nodes are fully connected to the $C$ nodes of the output layer of the AE-MLFLANN (Fig. \ref{fig_aemlflann}). The weights $W, b$ connecting the AE transformation layer to the output layer are learnt in the training phase.

\subsubsection{Learning}
In the final phase of the network, the last layer weights are updated. These are the connections between $A$ and $Y$ layers, which have been initialized randomly. In single-label FLANN literature, learning techniques like, particle swarm optimization (PSO) and backpropagation have been used thoroughly. Authors have experimented with the different configurations for multi-label classification \citep{law2019optimize}, and backpropagation seems to be quite fast and suitable. Thus, in AE-MLFLANN, backpropagation has been used as the learning mechanism to update the weights in the training phase.
First, the original input $X$ is sequentially transformed to $X'$, then $A$ in the transformation phase. Once transformation is done, the output in the last layer is computed as:
\begin{equation}\label{Eq_XML1}
  Y'_j=\phi \left\{ \sum A\cdot W_j \right\},
\end{equation}
where, $\phi$ is the activation function of the output layer.
Thus, for each data instance, a corresponding label-set is obtained as,
\begin{equation}
    Y_i^{'} = [ \ y_{i1}^{'}, \ y_{i2}^{'},...., \ y_{iC}^{'} \ ],\\
\end{equation}
If, $Y$ is the target label, the error, $E$, calculated using the original set of labels is,
\begin{align}
 \nonumber   E &= \ Y_i  \ -  \ Y_i^{'}\\
    &=  [ \ y_{i1} - y_{i1}^{'}, \ y_{i2} - y_{i2}^{'},...., \ y_{iC} - y_{iC}^{'} \ ]
\end{align}
The weights are updated using this error with the backpropagation technique. The change in weights $\Delta W^{t}$, which is a set of weight vectors $\Delta w_{ik}^{t}$ at $t^{th}$ iteration is given as,
\begin{align}
    \Delta w_{ik}^{t} =\mu \cdot a_{ik}^{t} \cdot\delta^{t}
\end{align}
$\mu$ is the learning rate. The change or gradient at $t^{th}$ iteration is given as,
\begin{align}
    \delta^{t} = Y^{'} \cdot (1 - Y^{'}) \cdot E
\end{align}
The connections weights, $W$ at the $(t+1)^{th}$ iteration is given by,
\begin{align}
    W^{t+1} = W^{t} + \Delta W^{t}
\end{align}
The weights are updated iteratively throughout the training phase.
Once the training is complete, the trained network can now predict multi-label outputs for any unknown data sample.

\subsection{Testing AE-MLFLANN}
In the testing phase, an unknown sample is fed to the AE-MLFLANN. This input is first functional expanded from $X_{test}$ to $X_{test}'$, then the autoencoder transformation-reduction is processed.
\begin{equation}\label{}
  A_{test}=\sigma_1\left\{ \sum W_1 \cdot X' + b_1 \right\},
\end{equation}
where, $\sigma_1$ is the activation function of the encoder.
At the final output the classification scores are predicted as,
\begin{equation}\label{}
 Y'_{test} = \phi\left\{ \sum W \cdot A + b\right\}.
\end{equation}
The scores are then converted to hard labels using a global threshold. This threshold has been set to $0.5$ as a midpoint of irrelevant class score $0$ and relevant class score $1$.
Thus, all classes with score higher than the threshold are marked as relevant, while the rest become irrelevant. In this way, the AE-MLFLANN is able to predict multi-label output for any unseen data.

\section{Experimental Analysis} \label{Sec_exp}
Two phases of experiments have been performed on the proposed model. The first and the more elaborate experimentation has been done on multi-label data, using the proposed AE-MLFLANN model, since it is the main focus of the paper. 
The second, more concise, set of experiments has been done using a single-label variation of the proposed model, named, AE-SLFLANN, to evaluate its performance on single-label data. As per the knowledge of the authors, this kind of two-layer network does not exist in literature, hence, the authors have tested its efficiency on both types of data.
Experiments have been performed with MATLAB 2017a on a Windows OS with Intel Core i7 processor and 16 GB RAM.

\subsection{AE-MLFLANN for Multi-label data}
The proposed AE-MLFLANN has been experimentally verified on five standard multi-label datasets from http://mulan.sourceforge.net/datasets-mlc.html and compared against six state-of-the-art multi-label classifiers. Emotions (music) \citep{trohidis2008multi}, flags (image) \citep{goncalves2013genetic}, scene (image) \citep{boutell2004learning}, CAL500 (music) \citep{turnbull2008semantic} and yeast (biology) \citep{elisseeff2002kernel} datasets have been used for the experiments. Eight standard multi-label performance metrics, namely, average precision (Avg Prec), Hamming loss (H Loss), one error, coverage, ranking loss (R Loss), micro F1, macro F1 and subset accuracy (Sub Ac) have been used here with 5-fold and 10-fold cross validation over the above mentioned datasets. These metrics include example-based metrics (H Loss and Sub Ac), ranking-based metrics (Avg Prec, one error, coverage and R Loss), and label-based metrics (macro F1 and micro F1).

\begin{table*}[htbp]
\centering
\caption{5-fold CV results}
\label{Tab_5CV}
\resizebox{\textwidth}{!}{
\begin{tabular}{llrrrrrrrr}
\textbf{Dataset}  & \textbf{Method}    & \textbf{Avg Prec} & \textbf{H Loss} & \textbf{One Error} & \textbf{Coverage} & \textbf{R Loss} & \textbf{Micro F1} & \textbf{Macro F1} & \textbf{Sub Ac} \\
                  & BR                 & 0.6190            & 0.2640          & 0.4470             & 0.4331            & 0.3270          & 0.5570            & 0.5650            & 0.1720          \\
                  & CC                 & 0.5790            & 0.2610          & 0.4200             & \textbf{0.4163}   & 0.2940          & 0.5730            & 0.5850            & 0.2210          \\
                  & ECC                & 0.6735            & 0.2409          & 0.4300             & 2.7559            & 0.6695          & 0.4761            & 0.4259            & 0.1517          \\
\textbf{Emotions} & ML-KNN             & 0.8101            & 0.1925          & 0.2581             & 1.7486            & 0.1565          & 0.6692            & 0.6321            & 0.2867          \\
                  & MLFLANN            & 0.7619            & 0.2381          & 0.3423             & 1.9678            & 0.1979          & 0.6188            & 0.6133            & 0.2124          \\

                  & BP-MLL             & 0.7982            & 0.2089          & 0.2969             & 1.7673            & 0.1622          & 0.6913            & 0.6624            & 0.2849          \\
                  & \textbf{AE-MLFLANN} & \textbf{0.8217}   & \textbf{0.1863} & \textbf{0.2395}    & 1.6861            & \textbf{0.1420} & \textbf{0.6925}   & \textbf{0.6769}   & \textbf{0.3221} \\
                  &                    &                   &                 &                    &                   &                 &                   &                   &                 \\
                  & BR                 & 0.7295            & 0.2764          & 0.2184             & 4.5837            & 0.4882          & 0.7153            & 0.6206            & 0.1547          \\
                  & CC                 & 0.7212            & 0.2830          & 0.2289             & 4.6103            & 0.4840          & 0.7050            & 0.6053            & 0.1800          \\
                  & ECC                & 0.6803            & 0.3212          & 0.2340             & 4.5574            & 0.6015          & 0.7217            & \textbf{0.6706}   & 0.1310          \\
\textbf{Flags}    & ML-KNN             & 0.8117            & 0.2924          & 0.2138             & 3.7480            & 0.2128          & 0.7007            & 0.5183            & 0.1497          \\
                  & MLFLANN            & 0.7696            & 0.3051          & 0.2707             & 4.0323            & 0.2739          & 0.6836            & 0.6188            & 0.1286          \\

                  & BP-MLL             & 0.8003            & 0.3247          & 0.2134             & 3.9796            & 0.2338          & 0.6904            & 0.5054            & 0.1256          \\
                  & \textbf{AE-MLFLANN} & \textbf{0.8235}   & \textbf{0.2656} & \textbf{0.1867}    & \textbf{3.7200}   & \textbf{0.2025} & \textbf{0.7219}   & 0.5802            & \textbf{0.1758} \\
                  &                    &                   &                 &                    &                   &                 &                   &                   &                 \\
                  & BR                 & 0.7075            & 0.1131          & 0.4483             & 1.2746            & 0.5593          & 0.5785            & 0.5700            & 0.4067          \\
                  & CC                 & 0.7092            & 0.1126          & 0.4437             & 1.2692            & 0.5546          & 0.5821            & 0.5748            & 0.4117          \\
                  & ECC                & 0.6648            & 0.1240          & 0.5056             & 1.4802            & 0.6440          & 0.5053            & 0.4879            & 0.3249          \\
\textbf{Scene}    & ML-KNN             & 0.8717            & 0.0839          & 0.2181             & 0.4504            & 0.0731          & 0.7438            & 0.7476            & 0.6386          \\
                  & MLFLANN            & 0.8451            & 0.1014          & 0.2534             & 0.5629            & 0.0942          & 0.7129            & 0.7250            & 0.5596          \\

                  & BP-MLL             & 0.7461            & 0.1608          & 0.4163             & 0.8687            & 0.1562          & 0.5738            & 0.5776            & 0.3947          \\
                  & \textbf{AE-MLFLANN} & \textbf{0.8752}   & \textbf{0.0761} & \textbf{0.2119}    & \textbf{0.4379}   & \textbf{0.0707} & \textbf{0.7562}   & \textbf{0.7702}   & \textbf{0.6398} \\
                  &                    &                   &                 &                    &                   &                 &                   &                   &                 \\
                  & BR                 & 0.2757            & 0.1414          & 0.3886             & 169.3485          & 0.7879          & 0.3085            & 0.0953            & 0.0000          \\
                  & CC                 & 0.2769            & 0.1388          & 0.3866             & 169.3507          & 0.7913          & 0.3078            & 0.0838            & 0.0000          \\
                  & ECC                & 0.2742            & \textbf{0.1381} & 0.3667             & 169.2636          & 0.7954          & 0.3040            & 0.0760            & 0.0000          \\
\textbf{CAL500}   & ML-KNN             & 0.4904            & 0.1491          & 0.1216             & 129.7710          & 0.1832          & 0.3112            & 0.0867            & 0.0000          \\
                  & MLFLANN            & 0.3391            & 0.2147          & 0.4462             & 166.5194          & 0.3636          & 0.3436            & 0.1311            & 0.0000          \\

                  & BP-MLL             & 0.4891            & 0.1543          & 0.1156             & \textbf{128.5593} & 0.1755          & 0.3669            & 0.1227            & 0.0000          \\
                  & \textbf{AE-MLFLANN} & \textbf{0.4996}   & 0.1467          & \textbf{0.1070}    & 130.1255          & \textbf{0.1623} & \textbf{0.3852}   & \textbf{0.1473}   & 0.0000          \\
                  &                    &                   &                 &                    &                   &                 &                   &                   &                 \\
                  & BR                 & 0.6692            & 0.1952          & 0.3136             & 9.0016            & 0.4622          & 0.6375            & 0.3695            & 0.1684          \\
                  & CC                 & 0.6710            & 0.1933          & 0.3049             & 8.9102            & 0.4576          & 0.6412            & 0.3723            & 0.1763          \\
                  & ECC                & 0.6684            & \textbf{0.1900} & 0.2685             & 8.9814            & 0.4646          & 0.6381            & 0.3523            & \textbf{0.1779} \\
\textbf{Yeast}    & ML-KNN             & 0.7436            & 0.1943          & 0.2346             & 6.6215            & 0.1871          & \textbf{0.6441}   & 0.3693            & 0.1771          \\
                  & MLFLANN            & 0.6454            & 0.3073          & 0.3583             & 8.3078            & 0.2891          & 0.5286            & 0.3590            & 0.0501          \\

                  & BP-MLL             & 0.7334            & 0.2312          & 0.2747             & 6.7463            & 0.1951          & 0.6322            & \textbf{0.4172}   & 0.1220          \\
                  & \textbf{AE-MLFLANN} & \textbf{0.7542}   & 0.2071          & \textbf{0.2338}    & \textbf{6.5337}   & \textbf{0.1766} & 0.6325            & 0.4034            & 0.1541
\end{tabular}}
\end{table*}

\begin{table}[htbp]
\centering
\caption{10-fold CV results}
\label{Tab_10CV}
\resizebox{\textwidth}{!}{
\begin{tabular}{llrrrrrrrr}
\textbf{Dataset}  & \textbf{Method}    & \textbf{Avg Prec} & \textbf{H Loss} & \textbf{One Error} & \textbf{Coverage} & \textbf{R Loss} & \textbf{Micro F1} & \textbf{Macro F1} & \textbf{Sub Ac} \\
                  & BR                 & 0.6911            & 0.2308          & 0.3693             & 2.7446            & 0.6177          & 0.5146            & 0.4713            & 0.1737          \\
                  & CC                 & 0.7019            & 0.2227          & 0.3609             & 2.7044            & 0.5887          & 0.5447            & 0.5032            & 0.1921          \\
                  & ECC                & 0.6759            & 0.2395          & 0.4298             & 2.7382            & 0.6619          & 0.4838            & 0.4302            & 0.1551          \\
\textbf{Emotions} & ML-KNN             & 0.8016            & \textbf{0.1906} & 0.2732             & 1.7773            & 0.1628          & 0.6703            & 0.6281            & 0.3070          \\
                  & MLFLANN            & 0.7599            & 0.2496          & 0.3172             & 2.0328            & 0.2051          & 0.6088            & 0.5965            & 0.1838          \\
                  & BP-MLL             & 0.7977            & 0.2159          & 0.2866             & 1.7722            & 0.1613          & \textbf{0.6818}   & \textbf{0.6681}   & 0.2511          \\
                  & \textbf{AE-MLFLANN} & \textbf{0.8156}   & 0.1962          & \textbf{0.2445}    & \textbf{1.7320}   & \textbf{0.1491} & 0.6747            & 0.6536            & \textbf{0.3068} \\
                  &                    &                   &                 &                    &                   &                 &                   &                   &                 \\
                  & BR                 & 0.7466            & 0.2685          & 0.2227             & 4.5250            & 0.4773          & 0.7248            & 0.6457            & 0.1711          \\
                  & CC                 & 0.7390            & 0.2639          & 0.2327             & 4.5042            & 0.4638          & 0.7247            & 0.6262            & 0.2174          \\
                  & ECC                & 0.6780            & 0.3265          & 0.2490             & 4.4987            & 0.6115          & 0.7241            & \textbf{0.6567}   & 0.1261          \\
\textbf{Flags}    & ML-KNN             & 0.8175            & 0.2764          & 0.2346             & 3.7221            & 0.2043          & 0.7242            & 0.5701            & 0.1145          \\
                  & MLFLANN            & 0.7693            & 0.3136          & 0.2702             & 4.0061            & 0.2763          & 0.6773            & 0.6000            & 0.1342          \\

                  & BP-MLL             & 0.7984            & 0.3225          & 0.2330             & 3.9437            & 0.2280          & 0.6975            & 0.5197            & 0.1258          \\
                  & \textbf{AE-MLFLANN} & \textbf{0.8235}   & \textbf{0.2656} & \textbf{0.1867}    & \textbf{3.7200}   & \textbf{0.2025} & \textbf{0.7249}   & 0.5802            & \textbf{0.1758} \\
                  &                    &                   &                 &                    &                   &                 &                   &                   &                 \\
                  & BR                 & 0.7175            & 0.1098          & 0.4375             & 1.2360            & 0.5402          & 0.5958            & 0.5903            & 0.4246          \\
                  & CC                 & 0.7190            & 0.1096          & 0.4350             & 1.2314            & 0.5370          & 0.5977            & 0.5928            & 0.4275          \\
                  & ECC                & 0.6719            & 0.1228          & 0.4981             & 1.4449            & 0.6329          & 0.5155            & 0.4977            & 0.3328          \\
\textbf{Scene}    & ML-KNN             & 0.8723            & 0.0893          & \textbf{0.2135}    & 0.4587            & 0.0743          & 0.7415            & 0.7448            & \textbf{0.6335} \\
                  & MLFLANN            & 0.8448            & 0.1015          & 0.2538             & 0.5650            & 0.0943          & 0.7132            & 0.7256            & 0.5525          \\

                  & BP-MLL             & 0.7361            & 0.1605          & 0.4261             & 0.8675            & 0.1652          & 0.5534            & 0.5876            & 0.4147          \\
                  & \textbf{AE-MLFLANN} & \textbf{0.8728}   & \textbf{0.0881} & 0.2164             & \textbf{0.4417}   & \textbf{0.0714} & \textbf{0.7506}   & \textbf{0.7659}   & 0.6215          \\
                  &                    &                   &                 &                    &                   &                 &                   &                   &                 \\
                  & BR                 & 0.2756            & 0.1426          & 0.3844             & 169.2574          & 0.7892          & 0.3061            & \textbf{0.1766}   & 0.0000          \\
                  & CC                 & 0.2790            & 0.1391          & 0.3844             & 169.2872          & 0.7911          & 0.3083            & 0.1710            & 0.0000          \\
                  & ECC                & 0.2756            & 0.1384          & 0.3705             & 169.2458          & 0.7977          & 0.3016            & 0.1617            & 0.0000          \\
\textbf{CAL500}   & ML-KNN             & 0.4910            & 0.1393          & 0.1174             & \textbf{130.5269} & 0.1894          & 0.3148            & 0.1734            & 0.0000          \\
                  & MLFLANN            & 0.3326            & 0.2181          & 0.4916             & 167.4366          & 0.3735          & \textbf{0.3379}   & 0.1745            & 0.0000          \\

                  & BP-MLL             & 0.4741            & 0.1633          & 0.1256             & 138.5593          & 0.1955          & 0.3369            & 0.1227            & 0.0000          \\
                  & \textbf{AE-MLFLANN} & \textbf{0.4984}   & \textbf{0.1373} & \textbf{0.1134}    & 139.8022          & \textbf{0.1888} & 0.3142            & \textbf{0.1766}   & 0.0000          \\
                  &                    &                   &                 &                    &                   &                 &                   &                   &                 \\
                  & BR                 & 0.6689            & 0.1962          & 0.3165             & 8.9787            & 0.4613          & 0.6366            & 0.3780            & 0.1746          \\
                  & CC                 & 0.6708            & 0.1944          & 0.3099             & 8.9183            & 0.4578          & \textbf{0.6400}   & 0.3773            & 0.1804          \\
                  & ECC                & 0.6693            & \textbf{0.1903} & 0.2751             & 8.9211            & 0.4621          & 0.6390            & 0.3615            & \textbf{0.1783} \\
\textbf{Yeast}    & ML-KNN             & 0.7518            & 0.1951          & 0.2356             & 6.3082            & 0.1795          & 0.6382            & 0.3768            & \textbf{0.1783} \\
                  & MLFLANN            & 0.6503            & 0.3032          & 0.3517             & 8.2581            & 0.2850          & 0.5335            & 0.4040            & 0.0567          \\

                  & BP-MLL             & 0.7326            & 0.2323          & 0.2714             & 6.7200            & 0.1945          & 0.6329            & \textbf{0.4180}   & 0.1109          \\
                  & \textbf{AE-MLFLANN} & \textbf{0.7545}   & 0.2032          & \textbf{0.2341}    & \textbf{6.5348}   & \textbf{0.1747} & 0.6360            & 0.3668            & 0.1485
\end{tabular}
}
\end{table}

Table \ref{Tab_5CV} and \ref{Tab_10CV} show the 5-fold and 10-fold cross validation results respectively for AE-MLFLANN on all datasets against the six other algorithms. These include three data transformation methods, BR, CC and ECC, and three problem adaptation techniques, ML-KNN, MLFLANN and BPMLL.
From a general overview, the proposed method is seen to perform substantially better than the other algorithms in comparison with all the eight performance metrics.
Delving deeper into specific comparisons, AE-MLFLANN is outright better than its single layer version, MLFLANN, for all the datasets and metrics. This indicates that the inclusion of the transformation layer has definitely proven to be fruitful. It is now able to perform multi-label classification more efficiently than before.
BP-MLL is a two-layer MLP model, which performs more weight adjustments and computations than AE-MLFLANN. However, AE-MLFLANN is seen to surpass BP-MLL for all the datasets in almost all aspects. This shows that the simple transformations of MLP may not be as efficient for multi-label data as our novel combination of functional expansion and feature transformation.
ML-KNN is a multi-label adaptation of KNN, which is computationally expensive and is seen to surpass AE-MLFLANN in rare cases. As a problem adaptation technique, AE-MLFLANN, establishes its performance quite well as compared to the above three methods.
A similar performance situation occurs for the three data transformation methods.
BR uses multiple classifiers (one for each class) but it is almost never seen to perform the best for any of the metrics.
CC also uses multiple classifiers, and performs very close to BR. However, very rarely does it outperform the proposed model.
The ensemble technique, ECC, performs better than BR and CC sometimes, but AE-MLFLANN surpasses it in almost all cases.

\begin{table}[htbp]
  \centering
  \caption{T-Test for all methods against AE-MLFLANN}\label{Tab_ttest}
  \begin{tabular}{lr}
    \textbf{Method} & \textbf{t-Test Value} \\
    BR              & 3.8329                \\
    CC              & 4.1265                \\
    ML-KNN          & 5.4046                \\
    MLFLANN         & 2.3138                \\
    ECC             & 3.4028                \\
    BP-MLL          & 2.0090
  \end{tabular}
\end{table}

Table \ref{Tab_ttest} depicts the t-test values for average precision of all the six methods against the proposed AE-MLFLANN. For $t_{.90}=1.533$ with degrees of freedom $ = 4$, AE-MLFLANN outperforms all the other methods.
Overall assessment of the results obtained show that the proposed AE-MLFLANN has proven to be quite efficient in the domain of multi-label classification.

\subsection{AE-SLFLANN for single-label data}
Although, AE-MLFLANN has been specifically developed for multi-label classification, its version AE-SLFLANN has been tested on single-label data as well to analyse its effectiveness.
\begin{table}[htbp]
\centering
\caption{Testing Accuracy for single-label data}
\label{Tab_SL}
\begin{tabular}{lrrrr}
\textbf{Datasets}       & \multicolumn{1}{r}{\textbf{MLP}} & \multicolumn{1}{c}{\textbf{FLANN}} & \multicolumn{1}{c}{\textbf{ELM}} & \multicolumn{1}{c}{\textbf{AE-SLFLANN}} \\
\textbf{Parkinson}      & 0.8215 & 0.8717 & 0.8042 & \textbf{0.8757} \\
\textbf{Ionosphere}     & 0.8632 & 0.9002 & 0.8776 & \textbf{0.9031} \\
\textbf{PIMA}           & 0.7721 & 0.7501 & 0.7383 & \textbf{0.7618} \\
\textbf{Vertebral 2C}   & 0.8323 & 0.8226 & 0.8419 & \textbf{0.8452}
\end{tabular}
\end{table}

Table \ref{Tab_SL} shows the testing accuracy of AE-SLFLANN on four datasets from the UCI repository. AE-SLFLANN works in a similar way as AE-MLFLANN, only the multi-label output layer at the end is replaced by a single-label one. The final classification is done by assigning hard label to the class with the maximum classification score, instead of using a threshold like in multi-label. Here, when compared to a single-label FLANN, the proposed AE-SLFLANN performs somewhat better than it. This happens, since the AE transformation used for AE-MLFLANN helps the multi-label data to be more separable, whereas, for AE-SLFLANN non-linearity introduced by the functional expansion was quite good, and the second layer contributes marginally but positively to it.
Additionally, AE-SLFLANN performs better than a two-layer MLP and an ELM in all cases. This strengthens the claim of the proposed network that the two consecutive layers performing expansion and transformation lead to an improved representation of the data which eventually leads to improved classification performance.

\section{Conclusion} \label{Sec_conc}
The proposed two-layer transformation network, which is an amalgamation of multi-label functional link artificial neural network and autoencoders is capable of efficient ML classification.
FLANN is a compact network that was originally used for single-label classification. However, its capability to transform data to a more separable space, made it a suitable candidate to be explored for complex multi-label data. Autoencoders on the other hand, are well-known for appropriate feature transformation and extraction.
The proposed network performs data transformation in two stages. The first layer functionally expands the original input, thus introducing non-linearity to it, while the second layer performs feature extraction and transformation through autoencoders to improve separability while reducing a bulk of the previously expanded input dimension. The final network is more compatible to handle multi-label data while transforming it to a further separable space. This network combines the strengths of FLANN and AE to overcome the bottlenecks of multi-label classification.
Experimental analysis of the proposed two-layer AE-MLFLANN has proven to surpass six standard multi-label classifiers for five datasets. Testing a single-label version of the proposed model also displays some improvement in comparison to three other algorithms on four single-label datasets. Overall, the proposed model is able to experimentally establish that the novel two-layer functional expansion and feature transformation is beneficial for both multi-label and single-label classification.
In future, this model can be expanded to a deeper framework with more transformations to handle more complex multi-label and single-label datasets.

\section*{Acknowledgement}
This work of Anwesha Law was supported by the Indian Statistical Institute, India.

\bibliography{References}
\end{document}